\documentclass[wcp]{jmlr}

\usepackage{longtable}

\usepackage{booktabs}

\usepackage{graphicx}
\usepackage{microtype}
\usepackage{graphicx}
\usepackage{booktabs} 
\usepackage{amsmath,amssymb} 
\usepackage{enumitem}
\usepackage{bm}
\usepackage{color}

\usepackage{hyperref}
\usepackage{arydshln}

\newcommand{\ie}{\textit{i}.\textit{e}., }
\newcommand{\eg}{\textit{e}.\textit{g}., }

\usepackage{lipsum}
\newcommand\blfootnote[1]{%
  \begingroup
  \renewcommand\thefootnote{}\footnote{#1}%
  \addtocounter{footnote}{-1}%
  \endgroup
}

\newif\ifjmlrutilssubfloats
\jmlrutilssubfloatstrue
\jmlrvolume{101}
\jmlryear{2019}
\jmlrworkshop{ACML 2019}

\title[Stochastic Branch]{Regularizing Neural Networks via Stochastic Branch Layers}

  \author{\Name{Wonpyo Park}\textsuperscript{1,4}*\Email{parkwonpyo@postech.ac.kr}\\
  \Name{Paul Hongsuck Seo}\textsuperscript{1,2}*\Email{hsseo@postech.ac.kr}\\
  \Name{Bohyung Han}\textsuperscript{2,3}\Email{bhhan@snu.ac.kr}\\
  \Name{Minsu Cho}\textsuperscript{1,3}\Email{mscho@postech.ac.kr}\\
  \addr \textsuperscript{1}Computer Vision Lab, CSE, POSTECH\\
        \textsuperscript{2}Computer Vision Lab, ECE \& ASRI, Seoul National University\\
        \textsuperscript{3}The Neural Processing Research Center\\
        \textsuperscript{4}Kakao Corporation
  }
\editors{Wee Sun Lee and Taiji Suzuki}

\begin{document}

\maketitle
\blfootnote{*Equal contribution}
\begin{abstract}
We introduce a novel stochastic regularization technique for deep neural networks, which decomposes a layer into multiple branches with different parameters and merges stochastically sampled combinations of the outputs from the branches during training.
Since the factorized branches can collapse into a single branch through a linear operation, inference requires no additional complexity compared to the ordinary layers.
The proposed regularization method, referred to as StochasticBranch, is applicable to any linear layers such as fully-connected or convolution layers.
The proposed regularizer allows the model to explore diverse regions of the model parameter space via multiple combinations of branches to find better local minima.
An extensive set of experiments shows that our method effectively regularizes networks and further improves the generalization performance when used together with other existing regularization techniques.
\end{abstract}


\section{Introduction}

Deep neural networks have made a remarkable progress in a variety of fields including computer vision, natural language processing, medical imaging, speech recognition, and computer graphics. 
Since many tasks in the fields require to understand high-level semantics, neural networks tend to go deeper and over-parametrized. 
Such deep and large networks are prone to overfitting so that a proper regularization becomes a critical factor in improving their generalization performance.
A popular type of regularization for deep neural networks is to inject random noise into the networks during training, \eg applying a binary random mask to hidden activations~\citep{hinton2012improving} or weights~\citep{wan2013regularization}, or skipping layers~\citep{huang2016deep} by forwarding activations via random identity connections.
Due to its simplicity and effectiveness, the stochastic regularization is widely used for training deep neural networks.

\begin{figure*}[t]
    \centering
    \hspace{0.2in}
    \includegraphics[width=0.8\linewidth]{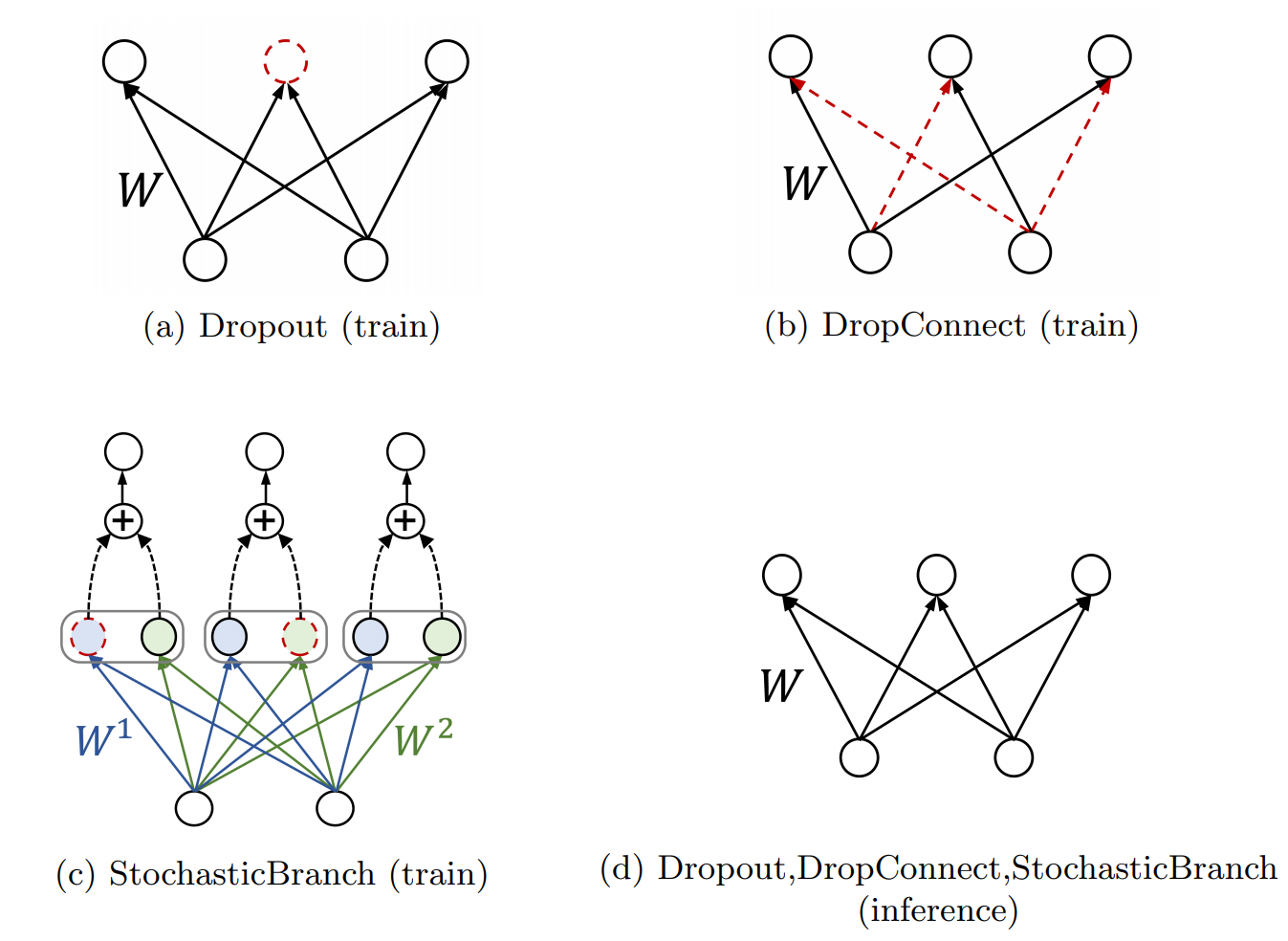}
    \caption{StochasticBranch compared to Dropout and DropConnect. These stochastic regularizers introduce the binary random mask during training. Red dashed lines represent what is masked out. Dropout and DropConnect apply random masks to activations and connections, respectively. StochasticBranch first decomposes each linear unit into multiple branches (two branches colored in blue and green) and injects random masks on the outputs of branches during training. At inference time, these branches are merge back to a single branch by taking the expectation of activations.
    Note that the inference procedures of Dropout, Dropconnect and StochasticBranch are exactly same and require the identical computational cost.
    StochasticBranch can be interpreted as the generalization of Dropout and DropConnect.}
    \label{fig:my_label}
\end{figure*}

We propose a novel regularization technique referred to as \textit{StochasticBranch}, which decomposes an ordinary linear layer into the one with multiple stochastic branches. 
By factorizing the original weight matrix of the layer into a set of matrices with their random binary masks, the stochastic branches effectively regularize a network during training. 
As a generalization of Dropout, its rich ensemble property with decomposed models allows to investigate exponentially many distinct models during training and explore diverse regions of the parameter space resulting in the better local optima. 
At inference time, the multiple branches collapse back into a single branch, thus requiring no additional complexity compared to the normal linear layers.
Fig.~\ref{fig:my_label} illustrates the comparison to Dropout~\citep{hinton2012improving} and Dropconnect~\citep{wan2013regularization}.
An extensive set of experiments shows the effectiveness of the proposed technique as well as wide applicability together with other popular regularizers including Dropout~\citep{hinton2012improving} and Batch Normalization~\citep{Ioffe:2015:BNA:3045118.3045167}.
%
%


\section{Related Work}

Regularization is a common and essential technique to combat overfitting in training. While one common form of the techniques is to penalize the weight tensor with a constant~\citep{NIPS1991_563,srebro2005rank}, a popular method for deep neural networks is to inject random noise during training. 
The most well-known example is Dropout~\citep{hinton2012improving} which stochastically zero out activations of neural networks to avoid co-adaptation of neurons.  Several successive follow-ups of Dropout have been proposed. 
\cite{wan2013regularization} propose a generalization of Dropout, called DropConnect, which zero-out weight rather than activation. 
\cite{ba2013adaptive} propose adaptive Dropout, where drop rate of activation is determined by a binary belief network overlaid on the neural network. 
\cite{li2016improved} introduce an efficient evolutionary Dropout that computes sampling probabilities on-the-fly from a mini-batch of examples. 
\cite{bulo2016dropout} develop Dropout distillation for better approximating the average predictor without sacrificing the computational efficiency of standard Dropout.
\cite{kang2016shakeout} propose Shakeout that randomly enhances or inverses contributions of each unit to the next layer, resulting in combination of L1 and L2 regularization.
\cite{gal2016dropout} introduce a theoretical framework for casting dropout as Bayesian inference  to approximate uncertainty of neural networks. 
\cite{zhai2018adaptive} propose a framework to adaptively adjust the drop rates based on the Rademacher complexity bound.


Other types of regularizers using stochastic noise have also been proposed to further improve generalization. 
\cite{zeiler2013stochastic} introduce a stochastic pooling that randomly picks activation within each pooling region according to a multinomial distribution. 
\cite{smith2016gradual} develop a dynamically growing neural network, Dropin, that gradually decreases the probability of skipping layers to train a network from shallow layers to deeper layers.
\cite{huang2016deep} propose to train a very deep ResNet network by stochastically skipping ResNet blocks via identity connections. 
\cite{ma2016dropout} introduce expectation-linear Dropout that regularizes the training objective with a measured inference gap. 
\cite{noh2017regularizing} reduce the gap between a marginal likelihood and a training objective with stochastic noise injection.

There exist recent methods that leverage multiple branches to improve generalization performance. 
\cite{lee2015m} propose a multi-head ensemble learning that shares early convolutional layers. \cite{han2017branchout} introduce a regularized ensemble method for single object tracking, which branches out intermediate layers to learn different target representations. Unlike our method, these approaches aim at proposing a specific type of architectures for ensemble learning rather than a generic regularization method for neural networks. 
\cite{goodfellow2013maxout} propose the Maxout activation function that merges outputs of branches of a single layer by max-pooling. It only helps the model with a stochastic regularizer better approximate ensemble results by model averaging, whereas our method itself is an effective stochastic regularizer.



\section{Stochastic Branch}
This section presents the details of StochasticBranch and relates the method to Dropout~\citep{hinton2012improving,srivastava2014dropout}.
For ease of explanation, we will only discuss a fully-connected (\texttt{fc}) layer in this section. 
Note, however, that StochasticBranch is applicable to any type of linear operators including convolution.

\subsection{Stochastic Branch Layer}
Let us consider an \texttt{fc} layer with input $\mathbf{x}\in\mathcal{R}^D$ and output $\mathbf{y}\in\mathcal{R}^{\hat{D}}$:  
\begin{equation}
    \mathbf{y}=\sigma\left(\mathbf{W}\mathbf{x}\right) ~~~~~\text{and}~~~~~ y_i=\sigma\left(\sum_{j=1}^D{w_{i,j}x_j}\right),  \label{eq:linear}
\end{equation}
where $\mathbf{W}$ is a weight matrix and $\sigma(\cdot)$ is an element-wise nonlinear activation function such as ReLU and tanh.

We decompose the weight matrix $\mathbf{W}$ into a sum of $N$ matrices, \ie, $\mathbf{W}=\sum_{k=1}^N\mathbf{W}^k$, such that the pre-activation for each output unit is given by the sum of $N$ linear projections: 
\begin{align}
    y_i = \sigma\left(\sum_{j=1}^D{w_{i,j}x_j}\right)
        = \sigma\left(\sum_{j=1}^{D} \sum_{k=1}^N {w_{i,j}^kx_j}\right). \label{eq:decomp_mtx}
\end{align}
This network structure can be interpreted as integrating multiple branches, where an output node is computed from the sum of $N$ branches.

In order to make the branches stochastic, we now introduce a random variable, $m_i^k\sim \mathrm{Bernoulli}(p^k)$, to each branch in Eq.~\eqref{eq:decomp_mtx}, thus resulting in
\begin{align}
    y_i = \sigma\left(\sum_{j=1}^{D}\sum_{k=1}^N m_i^k {w_{i,j}^k x_j}\right) 
    = \sigma\left(\sum_{k=1}^N{m_i^k\sum_{j=1}^{D}{w_{i,j}^k x_j}}\right). \label{eq:stochastic_branching}
\end{align}

As shown in Figure~\ref{fig:my_label}, each output node of the stochastic branch layer is obtained using $N$ stochastic branch units.
All input nodes of the layer are connected to the branch units, producing $N$ distinct values. The random binary masks then zero out a subset of the values, and
the corresponding output activation is computed from the sum of the masked values. 

Any linear layer parametrized by a weight matrix $\bm{W}$ can be transformed to a StochasticBranch layer with $N$ branches, \eg by setting the weight matrix of $k$-th branch to $\bm{W}^k$($k=1,...,N$) where the sum of those weights is $\bm{W}$. 
In training a neural network, the stochastic branch layers act as a regularizer. 
A set of $\hat{D}N$ random binary masks $\{ m_i^k \}$ is sampled for each training example and used in both forward and backward passes.
To update the weight matrices in the layer, \eg  via stochastic gradient descent (SGD), only the branches that were active in the forward pass are updated. Note that this stochastic training procedure induces the branches to be distinctive from each other. 
We present the effect in Section~\ref{exp}.


At inference time, instead of sampling random masks, we compute the output $y_i^*$ by taking the expectation of pre-activations using the following procedure:
\begin{align}
    y_i^* = \sigma\left(\mathbb{E}[y_i]\right) &= \sigma\left(\mathbb{E}\left[\sum_{k=1}^N{m_i^k\sum_{j=1}^{D}{w_{i,j}^kx_j}}\right]\right)  \nonumber \\
    &= \sigma\left(\sum_{k=1}^N{\mathbb{E}\left[m_i^k\right]\sum_{j=1}^{D}{w_{i,j}^kx_j}}\right)  \nonumber \\
    &= \sigma\left(\sum_{k=1}^N{p^k\sum_{j=1}^{D}{w_{i,j}^kx_j}}\right) \nonumber\\
    &= \sigma\left(\sum_{j=1}^D{\hat{w}_{i,j}x_j}\right),
\end{align}
where
\begin{align}
\hat{w}_{i,j}=\sum_{k=1}^N{p^k w_{i,j}^k}. \nonumber
\end{align}
Note that the multiple units branched for training now merge back into a single unit; there is no additional computational cost for inference compared to an ordinary \texttt{fc} layer as shown in Figure~\ref{fig:my_label}d.

\subsection{Generalized Dropout}
StochasticBranch is a generalization of Dropout~\citep{hinton2012improving,srivastava2014dropout} and DropConnect~\citep{wan2013regularization}, which can be shown below by imposing additional constraints on the StochasticBranch formulation.   

If we impose a group masking constraint that an identical mask $m_i$ is used for all branches with the same output unit $y_i$, \ie $m_i^k = m_i$ for all $k$, 
the multiple branches of Eq.~\eqref{eq:stochastic_branching} collapse into a single branch with a random mask variable:
\begin{align}
    y_i = \sigma\left(\sum_{k=1}^N{m_i\sum_{j=1}^{D}{w_{i,j}^kx_j}}\right) 
        = \sigma\left(m_i\sum_{j=1}^{D}{w_{i,j}x_j}\right).
\end{align}
For any zero-centered activation function $\sigma$ such as ReLU and tanh, we can move the mask variable $m_i$ out of the activation function so that it becomes equivalent to the Dropout regularizer~\citep{hinton2012improving,srivastava2014dropout}: 
\begin{equation}
    y_i = m_i\sigma\left(\sum_{j=1}^{D}{w_{i,j}x_j}\right) \label{eq:dropout}.
\end{equation}
This shows that Dropout is StochasticBranch under the constraint of group masks. Dropout either removes or retains an entire activation, whereas StochasticBranch rejects parts of the activation by masking out a subset of decomposed weights $\mathbf{W}^k$ in multiple branches.

If we impose a one-to-one branching constraint that each input $x_j$ is paired with exactly one branch, each input-output connection $w_{i,j}x_j$ involves a mask variable. For example, consider $D$ branches where $w^k_{ij}=w_{ij}$ for $k=j$ and  $w^k_{ij}=0$ for $k \neq j$. Then, StochasticBranch of Eq.~\eqref{eq:stochastic_branching} reduces to 
\begin{align}
    y_i = \sigma\left(\sum_{k=1}^D{m_i^k\sum_{j=1}^{D}{w_{i,j}^kx_j}}\right) 
        = \sigma\left(\sum_{k=1}^D{m_i^k{w_{i,k}x_k}}\right),
\end{align} 
which is exactly the same form with another generalized Dropout called DropConnect~\citep{wan2013regularization}: 
\begin{align}
    y_i = \sigma\left(\sum_{j=1}^D{m_{i,j}{w_{i,j}x_j}}\right).
\end{align}
This in turn shows that DropConnect is StochasticBranch under the constraint of one-to-one branches.  DropConnect~\citep{wan2013regularization} sample each input-output connection, whereas 
StochasticBranch maintains different weights across multiple branches and sample a branch, rather than a connection, for each unit. 



As a generalization of Dropout and DropConnect, 
StochasticBranch plays the role of a strong stochastic regularizer as will be discussed in the following subsection.  
Note, however, that a combination with other methods is also possible and may become a better regularizer as an extension. For example, if the random mask of Dropout is added to StochasticBranch, the combination can be represented as    
\begin{align}
    y_i = m_i\sigma\left(\sum_{k=1}^N{m_i^k\sum_{j=1}^{D}{w_{i,j}^kx_j}}\right). 
\end{align}
which is a further generalization of StochasticBranch with additional group masking $m_i$ of Dropout. 
Dropout is designed to mitigate the problem of co-adaptation that neurons excessively rely on other neurons~\citep{hinton2012improving,srivastava2014dropout}.  
By turning off neurons with probability of $1-p$, Dropout encourages neurons to less co-adapt each other and induces the layer to produce sparse activations.  
In contrast, StochasticBranch activates neurons via $2^N$ combinations of branch outputs, and induces the layer to produce diverse activations across examples. And, its turn-off chance is significantly smaller than that of Dropout, which is probability of $\prod_{k=1}^N(1-p^k)$.
Considering the differences, the two techniques may complement each other in practice. We will demonstrate such combination effects in Section~\ref{exp}.

\subsection{Discussion}
\label{sec:discussion}
\paragraph{Ensemble learning.} 
From an ensemble learning point of view~\citep{hinton2012improving,baldi2014dropout}, Dropout and its generalizations can be interpreted as learning an exponentially large ensemble of networks, where each model of the ensemble is given training examples in different orders via mini-batching during training. Each method approximates an ensemble from different classes of networks. 
Previous methods such as Dropout and DropConnect draw such models only within the original neural network. For example, Dropout approximates geometric ensemble averaging of $2^{\hat{D}}$ models~\citep{baldi2014dropout} where $\hat{D}$ denotes the number of units with Dropout. 
In contrast, StochasticBranch draws models from a richer class of networks augmented by branching so that the models in the ensemble are parameterized with different weights from distinct combinations of branches.
It thus approximates an ensemble of ${2^{\hat{D}N}}$ models where $\hat{D}$ and $N$ is the number of StochasticBranch units and branches, respectively. This rich ensemble with decomposed models allows to explore different regions of the parameter space and may find a diverse set of local optima. 



\paragraph{Data augmentation.} Dropout can also be seen as an implicit form of sophisticated data augmentation increasing training data coverage~\citep{konda2015dropout}, \eg in case of images, translations, rotations, scaling, etc.  Noise induced by a random mask $\mathbf{m}$ results in a similar effect of augmenting data $\mathbf{x}$ using a set of such transformations $f \in \mathcal{F}$ so that $\mathbf{m} \odot \sigma\left(\mathbf{W}\mathbf{x}\right) \approx \sigma\left(\mathbf{W}f(\mathbf{x})\right)$ in the case of a single layer model.
Here, a model in the Dropout ensemble can correspond to a transformation $f$ for data augmentation.
In this perspective, StochasticBranch creates a larger set of fine-grained transformations by decomposing transformations of Dropout, resulting in an effect of richer data augmentation. 



\paragraph{Batch normalization.} It has been widely known that Dropout and Batch Normalization are in disharmony each other in using them together. 
A recent research~\citep{li2018understanding} shows that a cause of the disharmony is a variance shift of activations in Dropout between training and testing, and suggests to reduce the variance shift by placing Batch Normalization before random noise injection of Dropout. 
For the same reason, when Batch Normalization is used together with StochasticBranch, we place Batch Normalization before noise injection of StochasticBranch.
Compared to Dropout, 
we observe that StochasticBranch has a lower variance shift, thus being more compatible with Batch Normalization. The corresponding experiments are reported in Section \ref{exp:mnist_and_fmnist}. 

\paragraph{Maxout.} Maxout networks~\citep{goodfellow2013maxout} also have a similar branching structure where a layer merges outputs of multiple branches. Despite its apparent similarity to StochasticBranch, its goal and structure are significantly different from ours.
Maxout is a non-linear activation function by max-pooling that is designed to improve ensemble approximation with stochastic regularizers by model averaging. 
It can thus also be used together with the proposed stochastic regularizer. Moreover, all branches of Maxout networks need to maintain their parameters even at inference due to the max-pooling operation, whereas multiple branches of StochasticBranch merge back into a single unit at inference.

\paragraph{Time Complexity.} 
SB introduces additional time complexity within only a few layers where SB is applied. Therefore, the increase of overall complexity is not significant in most cases. For instance, the use of SB on ResNet-110 in our experiments (refer to Table~\ref{tab:cifar100_resnet_mobilenet}) increases only 3.62\% of time complexity (~2.48G vs. ~2.57G flops). Note that SB increases time complexity only during training while the inference complexity remains the same as ordinary networks.

\begin{table*}[t]
\centering
\caption{Averaged classification error [\%] and standard deviation on MNIST and FMNIST with five runs.} 
\label{tab:mnist_result}
\scalebox{1}{
\begin{tabular}[b]{
@{~~}p{2cm}@{}|ccc|ccc}
			& \multicolumn{3}{c|}{MNIST (Error/stdev)} 		& \multicolumn{3}{c}{FMNIST (Error/stdev)}				\\ 
			& MLP3 	    & MLP5	    & CNN		    & MLP3        & MLP5        & CNN           \\ \hline
Baseline	& 1.79 \scriptsize{/ 0.06}   & 1.95 \scriptsize{/ 0.11}   & 0.88 \scriptsize{/ 0.02}   & 10.08 \scriptsize{/ 0.16}	& 10.12 \scriptsize{/ 0.16}  & 8.34 \scriptsize{/ 0.15}   \\
~+DO		& 1.46 \scriptsize{/ 0.03}	& 1.72 \scriptsize{/ 0.07}	& 0.68 \scriptsize{/ 0.03}	& 9.44 \scriptsize{/ 0.05}	& 9.96 \scriptsize{/ 0.19}   & 7.65 \scriptsize{/ 0.17}   \\
~+BN		& 1.54 \scriptsize{/ 0.06}   & 1.58 \scriptsize{/ 0.06}	& 0.74 \scriptsize{/ 0.05}	& 10.04	\scriptsize{/ 0.20}	& 9.68 \scriptsize{/ 0.16}   & 9.65 \scriptsize{/ 0.13}   \\
~+DO+BN		& 1.42 \scriptsize{/ 0.05}	& 1.42 \scriptsize{/ 0.04}	& 0.74 \scriptsize{/ 0.05}	& 9.37 \scriptsize{/ 0.12}	& 9.55 \scriptsize{/ 0.18}   & 9.05 \scriptsize{/ 0.13}   \\
~+SB		& 1.47 \scriptsize{/ 0.03}   & 1.55 \scriptsize{/ 0.05}   & 0.73 \scriptsize{/ 0.04}	& 9.60 \scriptsize{/ 0.14}	& 9.64 \scriptsize{/ 0.03}   & 8.04 \scriptsize{/ 0.17}   \\
~+SB+DO	    & 1.30 \scriptsize{/ 0.02}   & 1.34 \scriptsize{/ 0.03}   & 0.63 \scriptsize{/ 0.03}   & {\bf9.18} \scriptsize{/ 0.08}  & 9.52 \scriptsize{/ 0.04}   & 7.66 \scriptsize{/ 0.06}   \\
~+SB+BN 	& {\bf1.25} \scriptsize{/ 0.03}  & {\bf1.19} \scriptsize{/ 0.02}  & {\bf0.45} \scriptsize{/ 0.03}  & 9.25 \scriptsize{/ 0.09}   & {\bf9.19} \scriptsize{/ 0.16}  & {\bf7.36} \scriptsize{/ 0.20}  \\ \hline
\end{tabular}
}
\end{table*}

\section{Experiments}
\label{exp}

We evaluate StochasticBranch on multiple image classification benchmarks.
In the experiments, our method (SB) is compared with two of the most popular regularization methods: Dropout (DO) and Batch Normalization (BN). 
We set the drop rates of SB and DO to 0.5, and use 10 branches ($N=10$) unless specified otherwise.

\subsection{MNIST and Fashion-MNIST}
\label{exp:mnist_and_fmnist}
We first conduct a set of experiments on MNIST~\citep{lecun1998gradient} and Fashoin-MNIST (FMNIST)~\citep{xiao2017fashion}.
Both benchmarks consist of $28 \times 28$ grayscale images with $10$ class labels.
MNIST classes represent digits between 0 and 9, whereas FMNIST classes correspond to fashion items.
In both benchmarks, training and test sets contain 60,000 and 10,000 examples, respectively.
We test our method on multi-layer perceptrons (MLP) and convolutional neural networks (CNN).
We implement two MLPs with 3 and 5 layers (MLP3 and MLP5) where each intermediate layer has 1,024 hidden units.
CNN consists of two convolution (\texttt{conv}) layers with $5\times5$ kernels and two \texttt{fc} layers.
The output channels of the first and second \texttt{conv} layers are set as 32 and 64, respectively, while the number of hidden activations of the first \texttt{fc} layer is 1,024.
In every network, we use ReLU function for intermediate activations.
We train the three models without any regularization techniques as our baselines to reveal the improvements by regularization methods.
SB is applied to every layer of MLP3 and CNN.
For MLP5, we apply the technique to the first, third and fifth layers.
BN is applied to pre-activations of every layer and DO is placed in between every successive \texttt{fc} layers.


\begin{figure}
    \centering
    \includegraphics[width=0.55\linewidth]{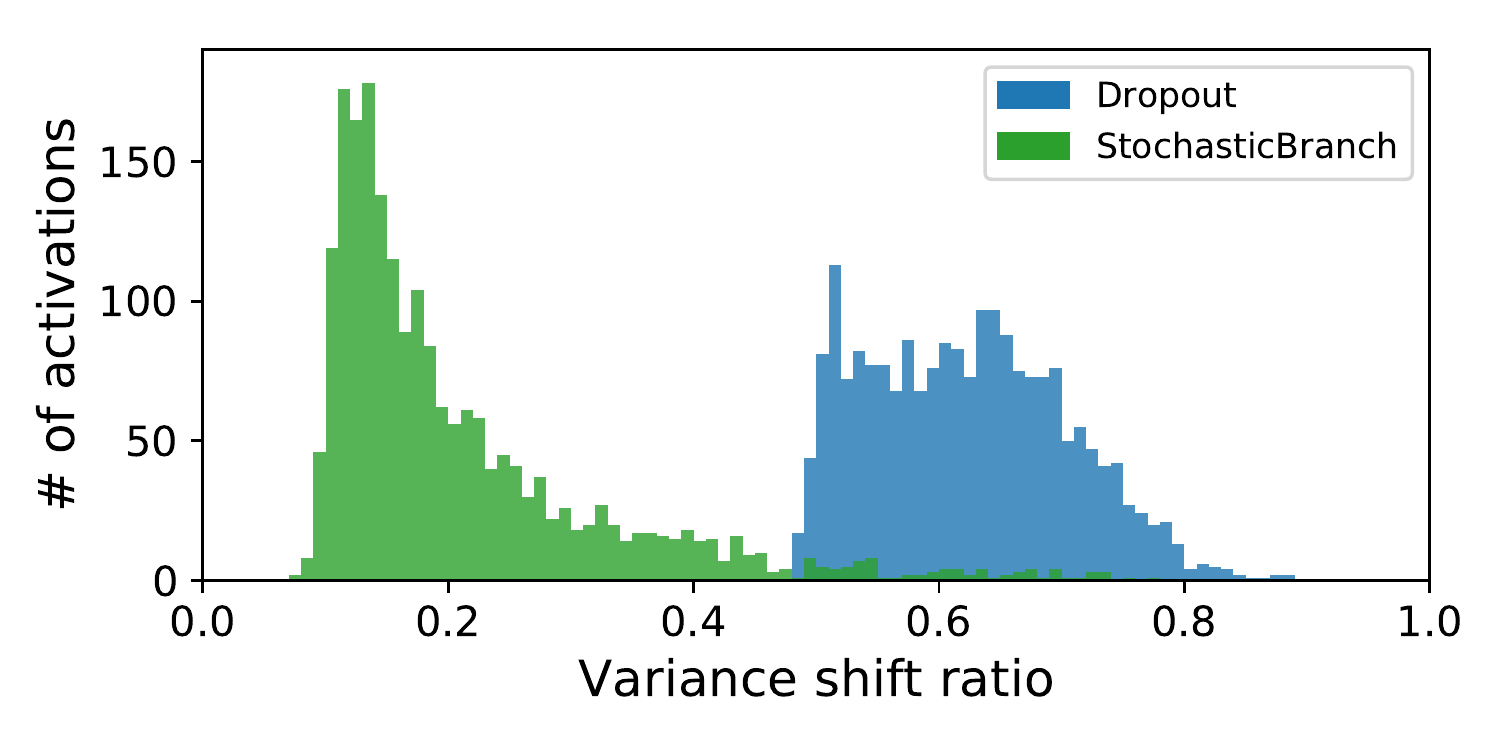}
    \caption{Histogram of variance shift ratio of activation after Dropout and StochasticBranch.}
    \vspace{-0.5cm}
    \label{fig:vsr_dosb}
\end{figure}

Table~\ref{tab:mnist_result} summarizes classification errors of the models with different regularization techniques on MNIST and FMNIST, where all results are obtained by averaging the errors of five independent runs.
The proposed method reduces the baseline errors in all settings comparable to or often better than other techniques.
Notably, our regularizer further reduces the errors when combined with other regularizers, achieving the largest error reductions in all settings.
%
%
Interestingly, Table~\ref{tab:mnist_result} reveals that the error reduction of SB with BN is significantly larger than that of DO with BN.
This is because SB has a lower variance shift compared to DO.
To quantify variance shifts of SB and DO, we measure variance shift ratio (VSR),which is given by
\begin{equation}
    \mathrm{VSR}=\frac{\mathrm{VAR}(y)-\mathrm{VAR}(y^*)}{\mathrm{VAR}(y)}
\end{equation}
where $\mathrm{VAR}(y)$ and $\mathrm{VAR}(y^*)$ represent the activation variances for each neuron at training and test time, respectively. 

Note that the ratio of 0 is the ideal case where there is no variance shift.
Figure~\ref{fig:vsr_dosb} presents VSR distribution of hidden activations of MLP3 on MNIST test set.
The figure clearly shows that activations of SB has lower variance shifts than DO.
In average, SB has VSR of 0.21 while DO shows that of 0.62.

In addition, we conduct experiments with Maxout to show that it is distinct from and complementary to StochasticBranch as discussed in Section~\ref{sec:discussion}.
We test DO and SB models of MLP3 with Maxout on MNIST by replacing the non-linearity function of the stochastic layers.
The use of Maxout further reduces the classification error by 0.04\% and 0.14\% for DO and SB, respectively.
Note that the additional error reduction by Maxout is larger with SB than with DO.

\begin{figure}[t]
    \includegraphics[width=\linewidth]{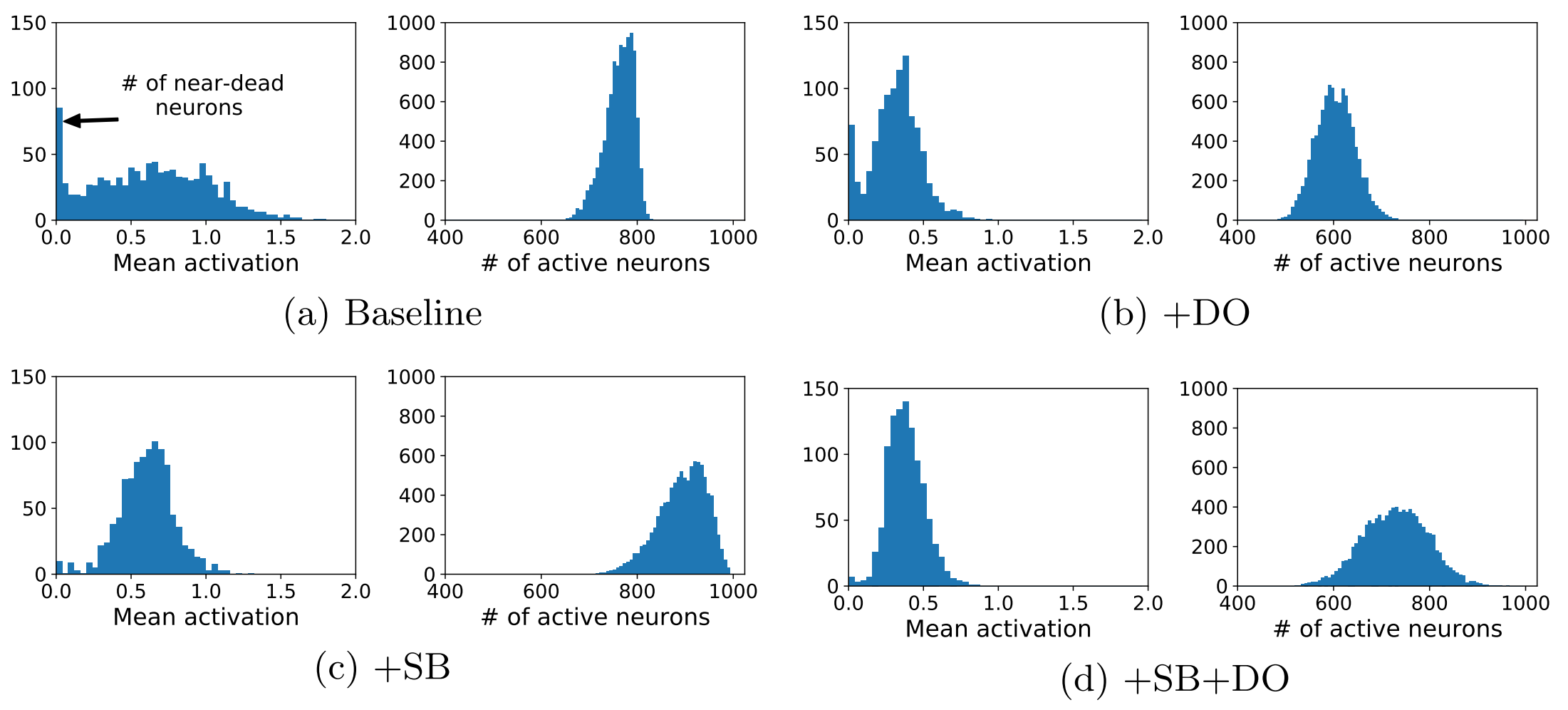}
    
    
    
    
    

    %
    \caption{Effects of StochasticBranch, Dropout, and StochasticBranch+Dropout on MNIST test set.     We present two histograms for each model: (left) the number of neurons vs. the mean activation value of the neuron over all images, (right) the number of images vs. the number of active neurons per image. 
    Note that a neuron with mean activation value 0 is a dead neuron, and also that an image with a small number of active neurons means a case with sparse activations.}
    \label{fig:sb_do}
\end{figure}

Figure~\ref{fig:sb_do} shows the effects of SB, DO, and SB+DO in terms of activation statistics on MNIST test set. 
The statistics are measured with the activations at the second \texttt{fc} layer of MLP3.
We present two histograms for each method:
(1) histogram of mean activation of each neuron and (2) histogram of the number of active neurons for each image.
A neuron with zero mean activation is a dead neuron, and an image with a small number of active neurons corresponds to a case with sparse
representations. 
Note that if the weights of a neuron converge to a point where its preactivation is severely biased to a negative value, the neuron may become  near-dead and almost never activates~\citep{maas2013rectifier}.
The comparison between the baseline (Figure~\ref{fig:sb_do}a) and DO (Figure~\ref{fig:sb_do}b) shows that DO reduces near-dead neurons as well as excessively-active neurons, and also induces sparse activations. The similar effects have been observed in the work of~\citep{srivastava2014dropout}. 
Interestingly, Figure~\ref{fig:sb_do}c shows that SB is significantly more effective in reducing near-dead neurons than DO. 
Since the process of stochastic branching generates exponentially many weight combinations, some combinations without negative bias may allow dead neurons to activate again by receiving the gradient during training.
Note that dead-neurons with ReLU is not able to be active again in both the baseline and DO since the dead-neurons never receive the gradient signal during training. 
As a side effect, the input features become denser as indicated by a large number of active neurons per image.
Due to these two different aspects (DO encouraging sparse representation and SB focusing more on reducing dead neurons), SB and DO may be complementary with each other.
Figure~\ref{fig:sb_do}d demonstrates that the combination of SB+DO lowers the number of active neurons per image and  retains sparse activations while reducing near-dead neurons. 

\begin{figure*}[t!]
    \centering
    \hspace{-0.3\textwidth}
    \includegraphics[width=\textwidth]{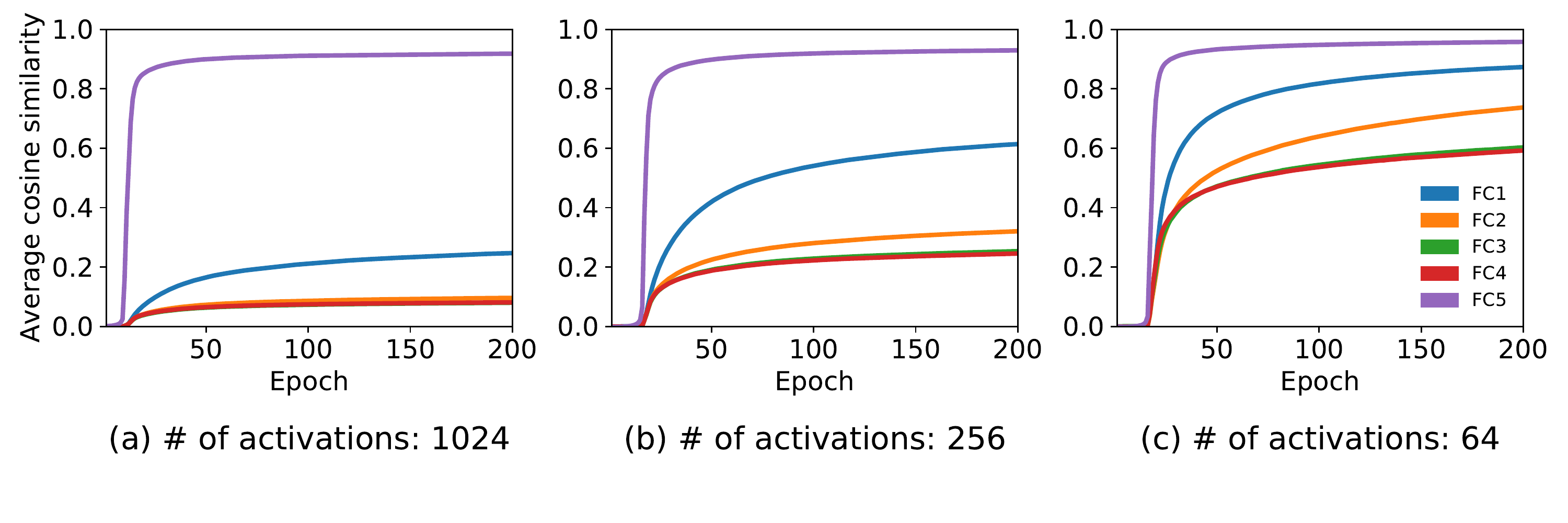}
    \hspace{-0.3\textwidth}
    \caption{
    (a-c) Average pairwise cosine similarity between branches varying number of hidden units in the network. 
    Cosine similarity of weight vectors for an activation in different branches is measured and averaged.
    We measure the similarity from every layer and observe the difference at different layers.
    }
    \label{fig:branch_cosine}
\end{figure*}

Figure~\ref{fig:branch_cosine} shows average cosine similarity of weight vectors between different branches, which is measured over epochs. 
For this experiment, we train three instances of MLP5 with different number of hidden units (64, 256, 1,024) applying SB to all layers. 
We observe significantly low cosine similarities between weights of branches 
in Figure~\ref{fig:branch_cosine}a;
this confirms that the branches of SB learn distinctive patterns and are capable of exploring various regions of the parameter space.
As we decrease the number of hidden units in the layers as in Figure~\ref{fig:branch_cosine}b and \ref{fig:branch_cosine}c, the branches become more similar because it is difficult to decompose a pattern with a small number of output units into diverse yet useful patterns.
\textcolor{black}{The redundant patterns across branches in these small networks bring lower regularization performance, resulting in relatively smaller gains.
For example, the accuracy gain (0.17\%) of the model with 64 activations is smaller than those (0.64\% and 0.35\%) of the models with 1024 and 256 activations. 
}
This implies that SB may perform better in layers with more output neurons.
Another observation is that the average similarity between branches tends to decrease when SB is applied to deeper layers.
The branches at the last layer (\texttt{fc5}) have particularly high similarity in all three settings since the last layer has only 10 output units corresponding to the number of classes.  


\subsection{CIFAR-10 and CIFAR-100}
\label{exp:cifar}
To validate the proposed method on more realistic settings, we conduct more experiments on CIFAR-10 and CIFAR-100~\citep{krizhevsky2009learning},  which contain $32\times32$ images with 10 and 100 object classes, respectively.
The sizes of training and test split are 50,000 and 10,000.
For experiments on CIFAR-10, we build a custom CNN that  consists of two $5\times5$ \texttt{conv}, one maxpool, two $5\times5$ \texttt{conv}, one maxpool, and two \texttt{fc} layers.
In this experiment, SB and DO are applied to \texttt{fc} layers. 
BN is applied to all \texttt{conv} layers as well as \texttt{fc} layers, which produces a larger performance gain of BN.
Table~\ref{tab:cifar10_error} summarizes the classification errors of the models on CIFAR-10, which show the similar tendency as on MNIST and FMNIST.
To study the effect of varying the number of branches, we also train another set of networks whose the first \texttt{fc} layers are  replaced by SB layers with 2, 4 or 8 branches.

\begin{figure}[t]
    \centering
    %
    %
     \includegraphics[width=0.9\linewidth]{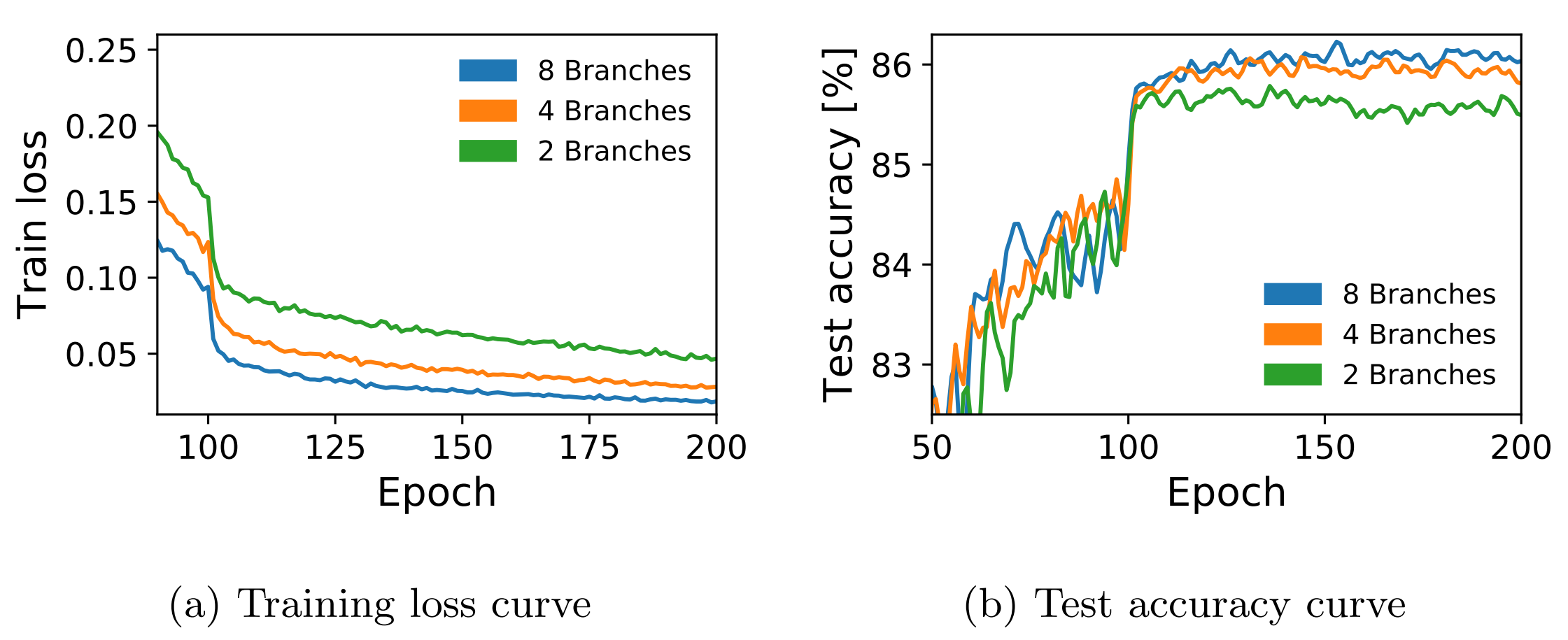}
    %
    \vspace{-0.1cm}
    \caption{Results on CIFAR-10:
    (a) Training loss curves and (b) test accuracy curves of models with different number of branches in SB layer.
    For clearance, we smooth test accuracy curves with window size of 5.
    }
    \label{fig:cifar10}
\end{figure}

We plot the curves of training loss and the test accuracy in Figure~\ref{fig:cifar10}a and \ref{fig:cifar10}b, respectively.
While SB injects stochastic noises diversifying hidden units, we notice that more branches help the network converge better compared to SB with fewer branches.
More branches create a richer ensemble of SB during training to better exploration of the parameter space, resulting in a higher chance to converge faster.

\begin{table}
    \centering
    \caption{Classification errors of models with different regularization techniques on CIFAR-10.}
    \label{tab:cifar10_error}
    \scalebox{1}{
    \begin{tabular}{
    @{~~}p{2.5cm}@{}|c
    }
    			& Error/stdev   \\ \hline
    Baseline	& 14.53 \scriptsize{/ 0.14}	        \\
    ~+DO		& 13.92 \scriptsize{/ 0.23}	        \\
    ~+BN		& 11.99 \scriptsize{/ 0.22}         \\
    ~+DO+BN		& 12.00 \scriptsize{/ 0.08}		    \\
    ~+SB		& 13.84 \scriptsize{/ 0.14}         \\
    ~+SB+DO	    & 13.82 \scriptsize{/ 0.24}         \\
    ~+SB+BN 	& {\bf11.83} \scriptsize{/ 0.29}    \\ \hline
    \end{tabular}
    }
\end{table}

\begin{table}
\centering
\caption{Classification error [\%] on CIFAR-10 with 1\% of training images.} 
\label{tab:cifar10_fewer}
\vspace{0.2cm}
\scalebox{1}{
\begin{tabular}[b]{
c|cccc
}
	& Baseline & +DO & +SB  & +SB+DO \\ \hline
Error	& 60.68 & 59.63 & 56.14 & \textbf{55.24}
\end{tabular}
}
\vspace{-0.1cm}
\end{table}

    

\vspace{-0.1cm}



To show the richer data augmentation effect of SB discussed in Section~\ref{sec:discussion}, we conduct additional experiments on CIFAR-10 with fewer training examples. 
We train models with 1\% of randomly sampled training examples for each class.
Table~\ref{tab:cifar10_fewer} shows the classification errors in this setting.
As shown in Table~\ref{tab:cifar10_fewer}, SB reduces the error more significantly than DO does where the relative error reduction of DO is 1.7\% whereas that of SB and SB+DO are 7.5\% and 9.0\%, respectively. 
These results imply that SB has a stronger data augmentation effect compared to DO.


For CIFAR-100, we train two advanced convolutional neural network architectures, ResNet-110~\citep{he2016deep} and MobileNetV2~\citep{sandler2018mobilenetv2}. 
For ResNet-110, both DO and SB are applied on the third \texttt{conv} layer of the last bottleneck block.
For MobileNetV2, both DO and SB are applied on the depthwise \texttt{conv} layer of the last two inverted residual blocks. 
The results are summarized in Table~\ref{tab:cifar100_resnet_mobilenet}. 
While DO performs similarly to the baseline, SB outperforms these models in both networks.
These results show the effectiveness of SB in \texttt{conv} layers.
It is worth noting that SB is more effective for layers with a large number of channels or a large kernel size. 
It is the reason why SB is used to the \texttt{conv} layers in the last blocks in these experiments.

\begin{table}
\vspace{-0.5cm}
\centering
\caption{
Classification error [\%] on CIFAR-100 tested with ResNet-110 and MobileNetV2 with single \texttt{fc}   layer. We apply regularization techniques on \texttt{conv} layers.}
\label{tab:cifar100_resnet_mobilenet}
\scalebox{1}{
\begin{tabular}{@{~~}p{4.5cm}@{}|c}
     & Error/stdev	\\ \hline
    ResNet-110 (Baseline) & 24.29 \scriptsize{/ 0.19} \\
    ResNet-110~+DO & 24.28 \scriptsize{/ 0.97} \\ 
    ResNet-110~+SB & \textbf{23.41} \scriptsize{/ 0.34} \\
    \hdashline
    MobileNetV2 (Baseline) & 27.82 \scriptsize{/ 0.39} \\
    MobileNetV2~+DO & 27.63 \scriptsize{/ 0.25} \\
    MobileNetV2~+SB & \textbf{27.16} \scriptsize{/ 0.32} \\ \hline
\end{tabular}
}
\end{table}


\subsection{PASCAL VOC}
In following experiments, we apply SB to pretrained networks .
To apply SB on a pretrained layer, We initialize the weight matrix of the branches by $\bm{W}^k=\frac{1}{Np^k}\bm{W}$ where $\bm{W}$ is the pretrained weight matrix.  
Note that although the weight matrices $\bm{W}^k$ are initialized by scaling the same pretrained weights $\bm{W}$, the stochastic branch layer is still capable of regularizing the network since each branch observes different training samples, and in consequence, the weight vectors of different branches become dissimilar with each other.

We conduct experiments on multiple tasks using PASCAL VOC~\citep{Everingham10} benchmarks: multi-label object classification, object detection, and semantic segmentation.
For evaluation metrics, mean average precision (mAP) is used for classification and detection while mean intersection over union (mIoU) is for semantic segmentation.
For all the tasks, ImageNet-pretrained VGG-16~\citep{DBLP:journals/corr/SimonyanZ14a} are used as our backbone networks, and SB is applied to \texttt{fc7} and DO is placed after the non-linear activations of \texttt{fc6} and \texttt{fc7}.

\begin{table}
\centering
\caption{Mean average precision [\%] of classification on PASCAL VOC 2012 validation and test sets.} 
\label{tab:voc_result_classification}
\vspace{-0.2cm}
\scalebox{1}{
\begin{tabular}{@{~~}p{2.1cm}@{}|cc}
     & Validation mAP/stdev	& Test mAP \\ \hline
Baseline    & 84.47 \scriptsize{/ 0.09} & 84.27 \\
~+DO        & 85.54 \scriptsize{/ 0.11} & 85.38 \\
~+SB        & 86.19 \scriptsize{/ 0.02} & 86.04 \\
~+SB+DO     & \textbf{86.50} \scriptsize{/ 0.03} & \textbf{86.36} \\ \hline
\end{tabular}
}
\end{table}


For object classification, we finetune the pretrained network on the training set after replacing the classification layer (\texttt{fc8}) to match the number of classes in PASCAL VOC; 
the trained models are evaluated on both validation and test sets.
Table~\ref{tab:voc_result_classification} shows mAPs of the multi-label classification models on PASCAL VOC 2012.
As seen in other experiments where we train networks from scratch, our method effectively improves the baseline performance by regularizing the network even though the weight vectors of branches are equally initialized.

\begin{table}[t]
\centering
\caption{Mean average precision [\%] of object detection on PASCAL VOC 2007 test set.} 
\label{tab:voc_result_objectdetection}
\vspace{0.3cm}
\scalebox{1}{
\begin{tabular}[b]{
c|ccc
}
			& Baseline                  & +DO                       & +SB                                          \\ \hline
mAP	& 70.17  & 70.36  & \textbf{71.41} 
\end{tabular}
}
\end{table}

\begin{table}[t]
\centering
\vspace{-0.2cm}
\caption{Mean intersection over union [\%] of semantic segmentation on PASCAL VOC 2011 validation set.} 
\vspace{-0.2cm}
\label{tab:voc_result_semanticsegmentation}
\scalebox{1}{
\begin{tabular}[b]{
c|ccc
}
			& Baseline                  & +DO                       & +SB                                          \\ \hline
mIoU	& 62.72 & 62.84  & \textbf{63.34} 
\end{tabular}
}
\end{table}
For object detection and semantic segmentation tasks, which require are more complex and structured prediction, we use Faster-RCNN~\citep{ren2015faster} and FCN-32S~\citep{long2015fully}, respectively, as the original architectures are equipped with DO as a regularizer.
Table~\ref{tab:voc_result_objectdetection} shows mAPs of the object detection models that are trained using training and validation sets of PASCAL VOC 2007 and evaluated on test set.
SB effectively regularizes the object detection network and outperforms both baseline and DO.
Note that Faster-RCNN architecture not only predicts class labels but also regresses bounding boxes.
Table~\ref{tab:voc_result_semanticsegmentation} presents the results of PASCAL VOC semantic segmentation task.
It also shows that SB is also effective for semantic segmentation resulting in an improved mIoU.

\label{sec:voc}


\section{Conclusion}
\label{conclusion}

In this paper, we proposed a novel regularization technique called StochasticBranch, which is a generalization of Dropout.
During training, the proposed method factorizes a single layer into multiple branches and sums up the outputs of the branches after random masking by binary noises.
At inference time, the multiple branches are merged back into a single layer.
We investigated that the proposed method regularizes the various neural networks on multiple benchmarks successfully and achieves significant improvement over the baseline performances.
Moreover, a set of experimental results show that our method can be applied with other commonly used regularizers such as Dropout or batch normalization achieving even further performance improvement.

\acks{This work is supported by Samsung Advanced Institute of Technology and by Basic Science Research Program (NRF-2017R1E1A1A01077999) through the National Research Foundation of Korea (NRF) funded by the Ministry of Science, ICT.}

\bibliography{acml19}

\begin{thebibliography}{33}
\providecommand{\natexlab}[1]{#1}
\providecommand{\url}[1]{\texttt{#1}}
\expandafter\ifx\csname urlstyle\endcsname\relax
  \providecommand{\doi}[1]{doi: #1}\else
  \providecommand{\doi}{doi: \begingroup \urlstyle{rm}\Url}\fi

\bibitem[Ba and Frey(2013)]{ba2013adaptive}
Jimmy Ba and Brendan Frey.
\newblock Adaptive dropout for training deep neural networks.
\newblock In \emph{Advances in Neural Information Processing Systems}, pages
  3084--3092, 2013.

\bibitem[Baldi and Sadowski(2014)]{baldi2014dropout}
Pierre Baldi and Peter Sadowski.
\newblock The dropout learning algorithm.
\newblock \emph{Artificial intelligence}, 210:\penalty0 78--122, 2014.

\bibitem[Bul{\`o} et~al.(2016)Bul{\`o}, Porzi, and
  Kontschieder]{bulo2016dropout}
Samuel~Rota Bul{\`o}, Lorenzo Porzi, and Peter Kontschieder.
\newblock Dropout distillation.
\newblock In \emph{International Conference on Machine Learning}, pages
  99--107, 2016.

\bibitem[Everingham et~al.(2010)Everingham, Van~Gool, Williams, Winn, and
  Zisserman]{Everingham10}
M.~Everingham, L.~Van~Gool, C.~K.~I. Williams, J.~Winn, and A.~Zisserman.
\newblock The pascal visual object classes (voc) challenge.
\newblock \emph{International Journal of Computer Vision}, 88\penalty0
  (2):\penalty0 303--338, June 2010.

\bibitem[Gal and Ghahramani(2016)]{gal2016dropout}
Yarin Gal and Zoubin Ghahramani.
\newblock Dropout as a bayesian approximation: Representing model uncertainty
  in deep learning.
\newblock In \emph{international conference on machine learning}, pages
  1050--1059, 2016.

\bibitem[Goodfellow et~al.(2013)Goodfellow, Warde-Farley, Mirza, Courville, and
  Bengio]{goodfellow2013maxout}
Ian~J Goodfellow, David Warde-Farley, Mehdi Mirza, Aaron Courville, and Yoshua
  Bengio.
\newblock Maxout networks.
\newblock \emph{arXiv preprint arXiv:1302.4389}, 2013.

\bibitem[Han et~al.(2017)Han, Sim, and Adam]{han2017branchout}
Bohyung Han, Jack Sim, and Hartwig Adam.
\newblock Branchout: Regularization for online ensemble tracking with
  convolutional neural networks.
\newblock In \emph{Proceedings of IEEE International Conference on Computer
  Vision}, pages 2217--2224, 2017.

\bibitem[He et~al.(2016)He, Zhang, Ren, and Sun]{he2016deep}
Kaiming He, Xiangyu Zhang, Shaoqing Ren, and Jian Sun.
\newblock Deep residual learning for image recognition.
\newblock In \emph{Proceedings of the IEEE conference on computer vision and
  pattern recognition}, pages 770--778, 2016.

\bibitem[Hinton et~al.(2012)Hinton, Srivastava, Krizhevsky, Sutskever, and
  Salakhutdinov]{hinton2012improving}
Geoffrey~E Hinton, Nitish Srivastava, Alex Krizhevsky, Ilya Sutskever, and
  Ruslan~R Salakhutdinov.
\newblock Improving neural networks by preventing co-adaptation of feature
  detectors.
\newblock \emph{arXiv preprint arXiv:1207.0580}, 2012.

\bibitem[Huang et~al.(2016)Huang, Sun, Liu, Sedra, and
  Weinberger]{huang2016deep}
Gao Huang, Yu~Sun, Zhuang Liu, Daniel Sedra, and Kilian~Q Weinberger.
\newblock Deep networks with stochastic depth.
\newblock In \emph{European Conference on Computer Vision}, pages 646--661.
  Springer, 2016.

\bibitem[Ioffe and Szegedy(2015)]{Ioffe:2015:BNA:3045118.3045167}
Sergey Ioffe and Christian Szegedy.
\newblock Batch normalization: Accelerating deep network training by reducing
  internal covariate shift.
\newblock In \emph{Proceedings of the 32Nd International Conference on
  International Conference on Machine Learning - Volume 37}, pages 448--456.
  JMLR.org, 2015.

\bibitem[Kang et~al.(2016)Kang, Li, and Tao]{kang2016shakeout}
Guoliang Kang, Jun Li, and Dacheng Tao.
\newblock Shakeout: A new regularized deep neural network training scheme.
\newblock In \emph{AAAI}, pages 1751--1757, 2016.

\bibitem[Konda et~al.(2015)Konda, Bouthillier, Memisevic, and
  Vincent]{konda2015dropout}
Kishore Konda, Xavier Bouthillier, Roland Memisevic, and Pascal Vincent.
\newblock Dropout as data augmentation.
\newblock \emph{stat}, 1050:\penalty0 29, 2015.

\bibitem[Krizhevsky and Hinton(2009)]{krizhevsky2009learning}
Alex Krizhevsky and Geoffrey Hinton.
\newblock Learning multiple layers of features from tiny images.
\newblock Technical report, Citeseer, 2009.

\bibitem[Krogh and Hertz(1992)]{NIPS1991_563}
Anders Krogh and John~A. Hertz.
\newblock A simple weight decay can improve generalization.
\newblock In J.~E. Moody, S.~J. Hanson, and R.~P. Lippmann, editors,
  \emph{Advances in Neural Information Processing Systems 4}, pages 950--957.
  Morgan-Kaufmann, 1992.

\bibitem[LeCun et~al.(1998)LeCun, Bottou, Bengio, and
  Haffner]{lecun1998gradient}
Yann LeCun, L{\'e}on Bottou, Yoshua Bengio, and Patrick Haffner.
\newblock Gradient-based learning applied to document recognition.
\newblock \emph{Proceedings of the IEEE}, 86\penalty0 (11):\penalty0
  2278--2324, 1998.

\bibitem[Lee et~al.(2015)Lee, Purushwalkam, Cogswell, Crandall, and
  Batra]{lee2015m}
Stefan Lee, Senthil Purushwalkam, Michael Cogswell, David Crandall, and Dhruv
  Batra.
\newblock Why m heads are better than one: Training a diverse ensemble of deep
  networks.
\newblock \emph{arXiv preprint arXiv:1511.06314}, 2015.

\bibitem[Li et~al.(2018)Li, Chen, Hu, and Yang]{li2018understanding}
Xiang Li, Shuo Chen, Xiaolin Hu, and Jian Yang.
\newblock Understanding the disharmony between dropout and batch normalization
  by variance shift.
\newblock \emph{arXiv preprint arXiv:1801.05134}, 2018.

\bibitem[Li et~al.(2016)Li, Gong, and Yang]{li2016improved}
Zhe Li, Boqing Gong, and Tianbao Yang.
\newblock Improved dropout for shallow and deep learning.
\newblock In \emph{Advances in Neural Information Processing Systems}, pages
  2523--2531, 2016.

\bibitem[Long et~al.(2015)Long, Shelhamer, and Darrell]{long2015fully}
Jonathan Long, Evan Shelhamer, and Trevor Darrell.
\newblock Fully convolutional networks for semantic segmentation.
\newblock In \emph{Proceedings of the IEEE conference on computer vision and
  pattern recognition}, pages 3431--3440, 2015.

\bibitem[Ma et~al.(2016)Ma, Gao, Hu, Yu, Deng, and Hovy]{ma2016dropout}
Xuezhe Ma, Yingkai Gao, Zhiting Hu, Yaoliang Yu, Yuntian Deng, and Eduard Hovy.
\newblock Dropout with expectation-linear regularization.
\newblock \emph{arXiv preprint arXiv:1609.08017}, 2016.

\bibitem[Maas et~al.(2013)Maas, Hannun, and Ng]{maas2013rectifier}
Andrew~L Maas, Awni~Y Hannun, and Andrew~Y Ng.
\newblock Rectifier nonlinearities improve neural network acoustic models.
\newblock In \emph{Proc. icml}, volume~30, page~2, 2013.

\bibitem[Noh et~al.(2017)Noh, You, Mun, and Han]{noh2017regularizing}
Hyeonwoo Noh, Tackgeun You, Jonghwan Mun, and Bohyung Han.
\newblock Regularizing deep neural networks by noise: Its interpretation and
  optimization.
\newblock In \emph{Advances in Neural Information Processing Systems}, pages
  5115--5124, 2017.

\bibitem[Ren et~al.(2015)Ren, He, Girshick, and Sun]{ren2015faster}
Shaoqing Ren, Kaiming He, Ross Girshick, and Jian Sun.
\newblock Faster r-cnn: Towards real-time object detection with region proposal
  networks.
\newblock In \emph{Advances in neural information processing systems}, pages
  91--99, 2015.

\bibitem[Sandler et~al.(2018)Sandler, Howard, Zhu, Zhmoginov, and
  Chen]{sandler2018mobilenetv2}
Mark Sandler, Andrew Howard, Menglong Zhu, Andrey Zhmoginov, and Liang-Chieh
  Chen.
\newblock Mobilenetv2: Inverted residuals and linear bottlenecks.
\newblock In \emph{2018 IEEE/CVF Conference on Computer Vision and Pattern
  Recognition}, pages 4510--4520. IEEE, 2018.

\bibitem[Simonyan and Zisserman(2014)]{DBLP:journals/corr/SimonyanZ14a}
Karen Simonyan and Andrew Zisserman.
\newblock Very deep convolutional networks for large-scale image recognition.
\newblock \emph{CoRR}, abs/1409.1556, 2014.

\bibitem[Smith et~al.(2016)Smith, Hand, and Doster]{smith2016gradual}
Leslie~N Smith, Emily~M Hand, and Timothy Doster.
\newblock Gradual dropin of layers to train very deep neural networks.
\newblock In \emph{Proceedings of the IEEE Conference on Computer Vision and
  Pattern Recognition}, pages 4763--4771, 2016.

\bibitem[Srebro and Shraibman(2005)]{srebro2005rank}
Nathan Srebro and Adi Shraibman.
\newblock Rank, trace-norm and max-norm.
\newblock In \emph{International Conference on Computational Learning Theory},
  pages 545--560. Springer, 2005.

\bibitem[Srivastava et~al.(2014)Srivastava, Hinton, Krizhevsky, Sutskever, and
  Salakhutdinov]{srivastava2014dropout}
Nitish Srivastava, Geoffrey Hinton, Alex Krizhevsky, Ilya Sutskever, and Ruslan
  Salakhutdinov.
\newblock Dropout: A simple way to prevent neural networks from overfitting.
\newblock \emph{The Journal of Machine Learning Research}, 15\penalty0
  (1):\penalty0 1929--1958, 2014.

\bibitem[Wan et~al.(2013)Wan, Zeiler, Zhang, Le~Cun, and
  Fergus]{wan2013regularization}
Li~Wan, Matthew Zeiler, Sixin Zhang, Yann Le~Cun, and Rob Fergus.
\newblock Regularization of neural networks using dropconnect.
\newblock In \emph{ICML}, pages 1058--1066, 2013.

\bibitem[Xiao et~al.(2017)Xiao, Rasul, and Vollgraf]{xiao2017fashion}
Han Xiao, Kashif Rasul, and Roland Vollgraf.
\newblock Fashion-mnist: a novel image dataset for benchmarking machine
  learning algorithms.
\newblock \emph{arXiv preprint arXiv:1708.07747}, 2017.

\bibitem[Zeiler and Fergus(2013)]{zeiler2013stochastic}
Matthew~D Zeiler and Rob Fergus.
\newblock Stochastic pooling for regularization of deep convolutional neural
  networks.
\newblock \emph{arXiv preprint arXiv:1301.3557}, 2013.

\bibitem[Zhai and Wang(2018)]{zhai2018adaptive}
Ke~Zhai and Huan Wang.
\newblock Adaptive dropout with rademacher complexity regularization.
\newblock In \emph{International Conference on Learning Representations}, 2018.

\end{thebibliography}

\appendix





\end{document}